\pdfoutput=1
\documentclass[sigconf]{acmart}
\usepackage{subcaption}
\usepackage{graphicx}
\usepackage{multirow}
\usepackage{pifont}
\usepackage{float}
\usepackage{stfloats}
\AtBeginDocument{%
  }

\setcopyright{acmlicensed}
\copyrightyear{2026}
\acmYear{2026}
\setcopyright{cc}
\setcctype{by}
\acmConference[MM '26]{Proceedings of the 34th ACM International Conference on Multimedia}{November 10--14, 2026}{Rio de Janeiro, Brazil}
\acmBooktitle{Proceedings of the 34th ACM International Conference on Multimedia (MM '26), November 10--14, 2026, Rio de Janeiro, Brazil}
\acmDOI{10.1145/3767308.3836108}
\acmISBN{979-8-4007-2213-4/2026/11}

\begin{document}

\title{SeCo-SBIR: Semantically Consistent Prompt Learning for Zero-Shot Sketch-Based Image Retrieval}

\author{Long Dang Hoang}
\affiliation{%
  \institution{Posts and Telecommunications Institute of Technology}
  \city{Ha Noi}
  \country{Vietnam}
}
\email{longdh@ptit.edu.vn}

\author{Tuan Nguyen Huu}
\affiliation{%
  \institution{Posts and Telecommunications Institute of Technology}
  \city{Ha Noi}
  \country{Vietnam}
}
\email{tuannh@ptit.edu.vn}

\author{Nguyen Minh Hieu}
\affiliation{%
  \institution{Posts and Telecommunications Institute of Technology}
  \city{Ha Noi}
  \country{Vietnam}
}
\email{hieunm.b23ce030@stu.ptit.edu.vn}

\author{Tu Minh Phuong}
\affiliation{%
  \institution{Posts and Telecommunications Institute of Technology}
  \city{Ha Noi}
  \country{Vietnam}
}
\email{phuongtm@ptit.edu.vn}

\renewcommand{\shortauthors}{Long et al.}

\begin{abstract}
Adapting CLIP for zero-shot sketch-based image retrieval (ZS-SBIR) via prompt learning faces a fundamental tension: the model must bridge the sketch-photo domain gap through task-specific adaptation, yet the added flexibility risks overfitting to seen training categories and eroding CLIP's zero-shot generalization.
We present \textbf{SeCo-SBIR}, a semantically consistent prompt learning framework that resolves this tension from both sides.
First, a \emph{text-guided multi-modal prompting} strategy routes learnable prompt vectors through CLIP's text encoder and projects the resulting intermediate representations into the visual encoder at every layer via learnable coupling functions. Because the text encoder has already learned robust, abstract category-level semantics from large-scale language supervision, this mechanism injects transferable semantic knowledge directly into the visual pathway - adapting the model to the sketch--photo domain while inherently favoring generalization to unseen classes.
Second, a \emph{perturbation-based consistency constraint} addresses the residual overfitting risk from the learnable coupling functions by aligning the adapted model with a frozen CLIP reference branch using an asymmetric InfoNCE objective - augmented inputs feed the frozen branch while clean inputs feed the trainable branch - anchoring the learned representations to CLIP's generalizable feature space.
Together with lightweight adapters and a multi-objective loss combining triplet, NT-Xent, and classification terms, SeCo-SBIR achieves state-of-the-art results on all three standard ZS-SBIR benchmarks across categorical, generalized, and across-dataset settings
\end{abstract}


\maketitle


\section{Introduction}

\begin{figure*}[t]
  \centering
  \resizebox{\textwidth}{!}{%
    \begin{subfigure}[t]{0.25\textwidth}
      \vspace{0pt}
      \centering
      \includegraphics[width=\textwidth]{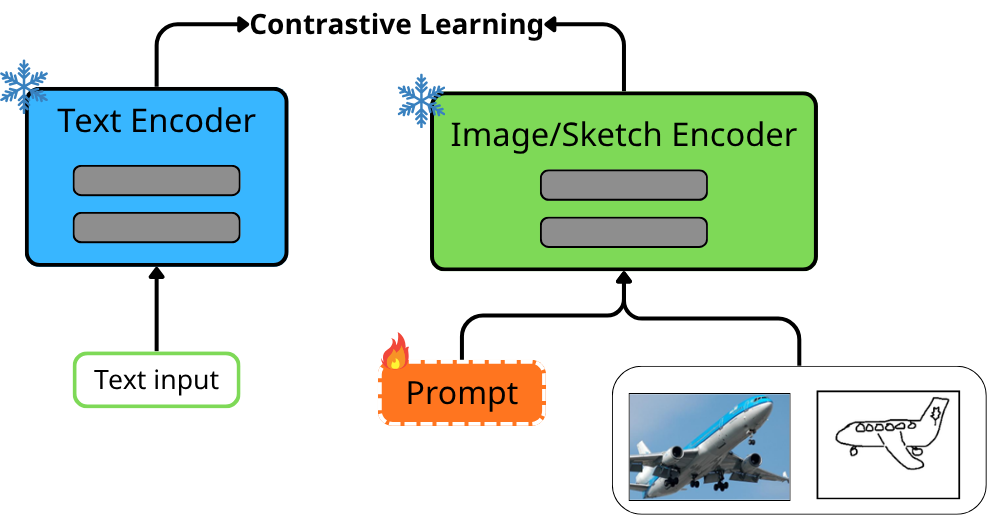}
      \vspace{0.2cm}
      \caption{CLIP-AT~\cite{sain2023clip}}
    \end{subfigure}
    \hfill
    \begin{subfigure}[t]{0.25\textwidth}
      \vspace{0pt}
      \centering
      \includegraphics[width=\textwidth]{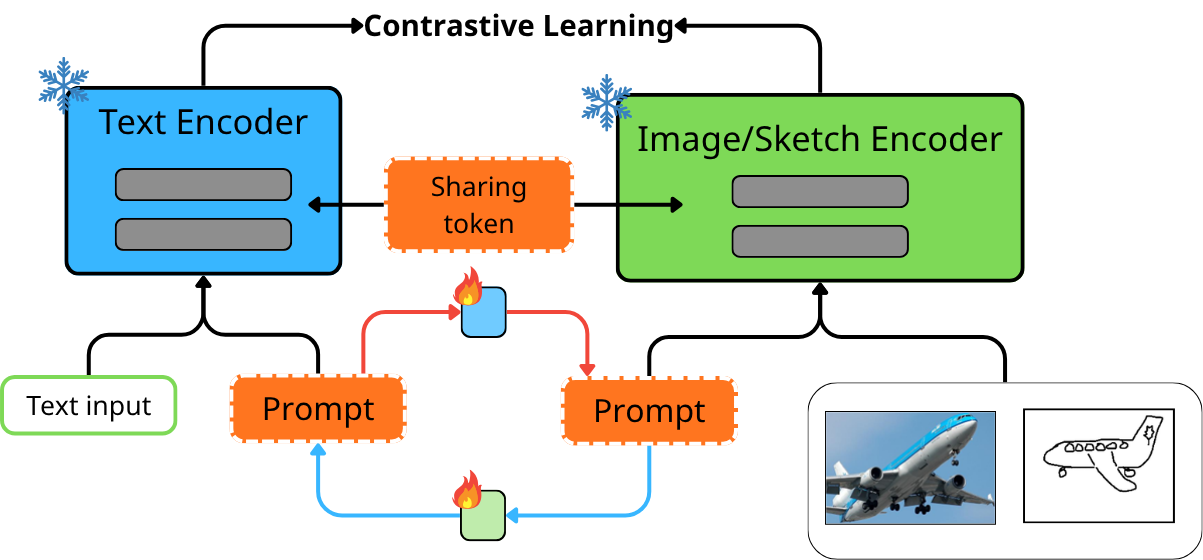}
      \vspace{0.3cm}
      \caption{SpLIP~\cite{singha2024splip}}
    \end{subfigure}
    \hfill
    \begin{subfigure}[t]{0.35\textwidth}
      \vspace{0pt}
      \centering
      \includegraphics[width=\textwidth]{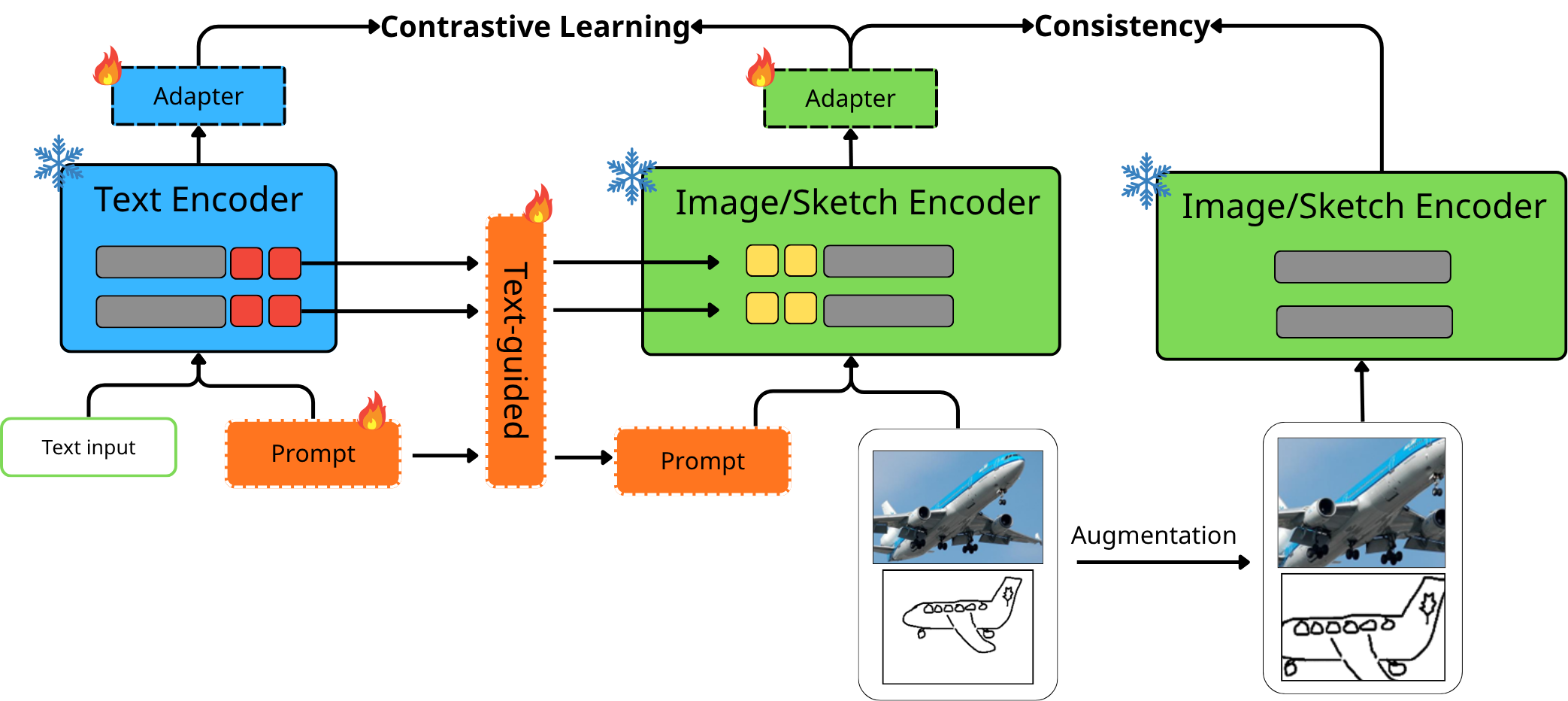}
      \caption{SeCo-SBIR (Ours)}
    \end{subfigure}%
  }
  \caption{Comparison of prompt learning paradigms for ZS-SBIR.
  (a)~CLIP-AT learns isolated visual prompts at the first encoder layer and uses fixed text templates, with no cross-modal prompt interaction.
  (b)~SpLIP introduces bidirectional token sharing between the text and visual encoders but lacks explicit regularization against seen-class overfitting.
  (c)~Our SeCo-SBIR routes learnable prompts through the text encoder and projects its intermediate representations into the visual encoder at every layer via learnable coupling functions (\emph{text-guided}), while a frozen reference branch with augmented inputs enforces a \emph{consistency} constraint that anchors the adapted embeddings to CLIP's generalizable feature space.
  Snowflake symbols (\ding{100}) denote frozen parameters; flame symbols denote trainable modules.}
  \label{fig:comparison}
\end{figure*}

Sketch-based image retrieval (SBIR) \cite{tuan2025unshare, yu2016sketch, song2016deep, liu2017deep} addresses the practical scenario in which a user draws a freehand sketch to query a database of natural images. Because sketches are produced with only a few strokes, they are inherently abstract and lack the rich texture, color, and fine detail of photographs, creating a significant cross-modal gap that any SBIR system must bridge. In the more challenging zero-shot setting (ZS-SBIR)~\cite{shen2018zero, yelamarthi2018zero, dutta2019semantically, dey2019doodle, liu2019semantic, jing2022augmented, sain2022sketch3t}, the model is trained on one set of categories and evaluated on entirely unseen ones, demanding not only cross-modal alignment but also robust category-level generalization. Generalized ZS-SBIR (GZS-SBIR)~\cite{dutta2019semantically} raises the difficulty further by mixing seen- and unseen-class photos in the retrieval gallery, which biases the model toward retrieving familiar categories and penalizes overfitting to the training distribution.

Vision-language foundation models, particularly CLIP~\cite{radford2021learning}, have become the dominant backbone for ZS-SBIR~\cite{sain2023clip, singha2024splip, lyou2024modality, li2024drclip, zhou2024mcg}, as its pre-trained dual-encoder architecture provides a rich semantic space that naturally accommodates the sketch, photo, and text modalities. The prevailing adaptation strategy is prompt learning~\cite{jia2022vpt, zhou2022coop, zhou2022cocoop}, which keeps most parameters frozen and introduces learnable tokens to steer the model toward the target domain. Yet existing prompt-based ZS-SBIR methods differ in how they leverage CLIP's multi-modal structure, and each leaves gaps that limit generalization.
CLIP-AT~\cite{sain2023clip}, the first to adapt CLIP for ZS-SBIR, injects shallow visual prompts at the first transformer layer while relying on fixed text templates, treating the two branches in isolation. SpLIP~\cite{singha2024splip} advances this with a bidirectional prompt exchange—mapping visual patches into textual tokens and projecting class-aggregated text outputs back into layer-specific visual prompts. While effective, two limitations persist: the visual prompts blend all seen-category semantics into a single aggregated signal without per-category grounding, and no explicit regularization prevents the adapted representations from drifting toward seen-class distributions. Other recent methods address complementary aspects - MARL~\cite{lyou2024modality} disentangles modality-specific information, Dr.CLIP~\cite{li2024drclip} refines retrieval features, MCG~\cite{zhou2024mcg} introduces modality capacity guidance, DCDL~\cite{li2025dcdl} applies causal disentangled learning, and SketchFusion~\cite{koley2025sketchfusion} fuses multiple foundation models - yet none explicitly addresses the core tension: \emph{the model must adapt deeply enough to bridge the sketch-photo domain gap, yet any additional flexibility risks overfitting to the seen training classes, undermining the very zero-shot generalization that CLIP provides}.

In this work, we resolve this tension with \textbf{SeCo-SBIR}, a \emph{semantically consistent prompt learning} framework for ZS-SBIR built on two complementary ideas. As illustrated in Fig.~\ref{fig:comparison}, our design departs from both the isolated prompting of CLIP-AT and the bidirectional exchange of SpLIP by addressing both sides of the tension: we strengthen adaptation through semantics that are inherently generalizable, and we add a principled regularizer that prevents the remaining learnable components from drifting toward seen classes.

\textbf{(i) Text-guided multi-modal prompting.} The key observation is that CLIP's text encoder has already learned robust, abstract category-level semantics from large-scale language supervision - knowledge that naturally transfers across categories because it is grounded in language rather than in the visual patterns of any particular training class. Rather than learning visual prompts in isolation from language~\cite{sain2023clip} or relying on bidirectional token exchange~\cite{singha2024splip}, we route learnable prompt vectors through the text encoder and project its intermediate representations into the visual encoder at every layer via \emph{learnable coupling functions}. This injects pre-existing transferable semantic knowledge directly into the visual pathway, producing layer-specific visual prompts whose category-level grounding comes from language rather than from seen-class visual statistics. The result is a deep adaptation mechanism that narrows the sketch-photo domain gap while inherently favoring generalization to unseen classes.

\textbf{(ii) Perturbation-based consistency constraint.} While text-guided prompting provides an adaptation signal that is semantically grounded and transferable, the learnable coupling functions and prompt vectors still introduce additional model capacity that risks overfitting to seen categories - a concern no prior ZS-SBIR method explicitly addresses. We counteract this residual risk by aligning the adapted embeddings with those of a frozen CLIP reference branch through an asymmetric InfoNCE objective: augmented inputs feed the frozen branch while clean inputs feed the trainable branch, preventing trivial copying and anchoring the learned representations to CLIP's generalizable feature space.

Combined with lightweight adapters and a multi-objective loss (triplet, NT-Xent, classification), SeCo-SBIR achieves state-of-the-art results on all three standard benchmarks. On TU-Berlin-Ext, the benchmark most sensitive to seen-class overfitting, we surpass the previous best by {+5.6\%} mAP@all (ZS-SBIR) and {+4.6\%} mAP@all (GZS-SBIR). In the across-dataset setting - training on Sketchy-Ext and evaluating on QuickDraw-Ext - we improve mAP@all by {+6.8\%}, underscoring the cross-domain robustness of our semantically consistent design.

In summary, our contributions are as follows:
\begin{itemize}
    \item We propose SeCo-SBIR, a prompt learning framework for ZS-SBIR that introduces a \emph{text-guided multi-modal prompting} mechanism with a learnable coupling function, which derives semantically grounded visual prompts from the text encoder's intermediate representations at every layer. By leveraging the text encoder's pre-existing, abstract category-level semantics, this mechanism adapts the visual pathway through knowledge that is inherently transferable to unseen classes.
    \item We introduce a \emph{perturbation-based consistency constraint} that aligns the adapted model with a frozen CLIP reference via an asymmetric InfoNCE objective, effectively regularizing the additional flexibility of the learnable coupling and preventing seen-class overfitting.
    \item We achieve state-of-the-art results across all three standard ZS-SBIR benchmarks in the categorical, generalized, and across-dataset evaluation settings, with especially pronounced gains where overfitting and domain shift are most severe.
\end{itemize}
\section{Related Work}

\subsection{Zero-Shot Sketch-Based Image Retrieval}

ZS-SBIR~\cite{yelamarthi2018zero, tian2021relationship, du2025zeroshot, tian2022tvt, wang2022prototype} requires retrieving photos matching a query sketch from unseen categories, demanding both cross-modal alignment and category-level generalization. Early methods employed ConvNet backbones with auxiliary semantic embeddings~\cite{dey2019doodle, dutta2019semantically, liu2019semantic, yelamarthi2018zero}, while later works introduced graph convolutional networks~\cite{zhang2020zeroshot}, Vision Transformers~\cite{tian2022tvt}, and hybrid architectures~\cite{lin2023zeroshot, gupta2022zeroshot} to learn more unbiased shared feature spaces. The generalized setting (GZS-SBIR)~\cite{dutta2019semantically} further tests robustness when seen- and unseen-class photos coexist in the gallery.

CLIP~\cite{radford2021learning} has reshaped ZS-SBIR. CLIP-AT~\cite{sain2023clip} first adapted CLIP using shallow visual prompts and fixed text templates. Subsequent methods improve the embedding space from different angles: MARL~\cite{lyou2024modality} disentangles modality-specific information, Dr.CLIP~\cite{li2024drclip} refines retrieval features, MCG~\cite{zhou2024mcg} balances sketch and photo representations via modality capacity guidance, SpLIP~\cite{singha2024splip} introduces bidirectional multi-modal prompt sharing, DCDL~\cite{li2025dcdl} applies causal disentangled learning, and SketchFusion~\cite{koley2025sketchfusion} fuses multiple foundation models. Despite steady progress, these methods focus on building more discriminative cross-modal embeddings without explicitly addressing the risk that deeper adaptation may cause overfitting to seen categories - the central challenge our work targets.

\subsection{Prompt Learning for VLMs}

Prompt learning adapts VLMs \cite{yao2023kgcoop, yao2024tcp, yang2023mma} by learning a small set of tokens while keeping most parameters frozen. CoOp~\cite{zhou2022coop} and CoCoOp~\cite{zhou2022cocoop} learn textual context vectors, while VPT~\cite{jia2022vpt} injects learnable tokens into the visual encoder. Multi-modal approaches go further: MaPLe~\cite{khattak2023maple} couples text and visual prompts via a fixed projection at each layer, and SpLIP~\cite{singha2024splip} extends this to ZS-SBIR with bidirectional token exchange between both encoders. Other works explore prompt-based domain generalization~\cite{bose2024stylip, singha2024unknown} and language-aware optimization~\cite{bulat2023lasp}.

A related line of work tackles the generalization--adaptation trade-off: PromptSRC~\cite{khattak2023promptsrc} regularizes prompted features against frozen CLIP outputs, and ConsistencyGPL~\cite{roy2024consistency} enforces consistency with the pre-trained backbone. However, these methods target image classification where the domain gap is modest. In ZS-SBIR, the sketch-photo gap is far more severe, requiring both deeper adaptation and stronger regularization simultaneously. SeCo-SBIR addresses this dual need through text-guided multi-modal prompting for semantically grounded adaptation and a perturbation-based consistency constraint to preserve generalization.

\section{Preliminaries: CLIP}
Contrastive Language-Image Pre-training (CLIP)~\cite{radford2021learning, rao2022denseclip} jointly trains a visual encoder~$\mathcal{V}$ and a text encoder~$\mathcal{T}$ on large-scale image-text pairs. CLIP supports both ResNet~\cite{he2016deep} and ViT~\cite{dosovitskiy2021image} backbones; following recent ZS-SBIR works~\cite{sain2023clip, singha2024splip}, we adopt the ViT variant.
 
\paragraph{Visual encoder.}
An input image $x \in \mathbb{R}^{H \times W \times 3}$ is split into $N$ patches, linearly embedded as
$E_0 = \{e^j_0\}_{j=1}^{N}$ with $e^j_0 \in \mathbb{R}^{d_v}$,
and prepended with a learnable \texttt{[CLS]} token
$c^v_0 \in \mathbb{R}^{d_v}$.
The sequence $[c^v_0, E_0]$ passes through $L$ transformer layers,
and a projection head maps the final \texttt{[CLS]}
representation to a $d$-dimensional joint embedding
$f_v = \mathcal{V}(x) \in \mathbb{R}^{d}$.
 
\paragraph{Text encoder.}
A sentence with $n$ tokens is converted into word embeddings
$W_0 = \{w^j_0\}_{j=1}^{n}$ with $w^j_0 \in \mathbb{R}^{d_t}$,
prepended with a \texttt{[CLS]} token $c^t_0 \in \mathbb{R}^{d_t}$,
and processed through transformer layers.
A projection head on the final \texttt{[CLS]} output yields the
textual feature $f_t = \mathcal{T}(\mathcal{S}) \in \mathbb{R}^{d}$.
 
\paragraph{Zero-shot classification.}
CLIP constructs textual prompts such as \texttt{"a photo of a [category]"}
for each of $N_c$ categories, obtaining class-specific features
$\{f^k_t\}_{k=1}^{N_c}$, and predicts:
\begin{equation}\label{eq:clip-pred}
  \mathcal{P}(y \mid x)
  = \frac{\exp\!\bigl(\mathrm{sim}(f_v,\, f^{y}_t) / \tau\bigr)}
         {\sum_{k=1}^{N_c} \exp\!\bigl(\mathrm{sim}(f_v,\, f^{k}_t) / \tau\bigr)},
\end{equation}
where $\mathrm{sim}(\cdot,\cdot)$ denotes cosine similarity
and $\tau$ is a learned temperature.

\section{Methodology}
\subsection{Problem Formulation}
In sketch-based image retrieval (SBIR), the goal is to retrieve a set of
$K$ photos $\{p_k\}_{k=1}^{K} \in \mathcal{P}$ from a gallery~$\mathcal{G}$,
given a query sketch $s \in \mathcal{S}$ belonging to a specific category
from a total of $C$ classes.
In the zero-shot setting (ZS-SBIR), $C$ is partitioned into
seen training classes~$\mathcal{C}^s$ and unseen testing classes~$\mathcal{C}^u$,
where $C = \mathcal{C}^s \cup \mathcal{C}^u$ and $\mathcal{C}^s \cap \mathcal{C}^u = \emptyset$.
The training set
$\mathcal{G}^s = (\mathcal{S}^s, \mathcal{P}^s, \mathcal{C}^s)$
consists of sketches~$\mathcal{S}^s$ and photos~$\mathcal{P}^s$
drawn from $\mathcal{C}^s$ categories,
while the test set
$\mathcal{G}^u = (\mathcal{S}^u, \mathcal{P}^u, \mathcal{C}^u)$
contains sketches and photos from unseen categories~$\mathcal{C}^u$.
A more challenging variant, generalized ZS-SBIR (GZS-SBIR)~\cite{dutta2019semantically},
constructs the test-time gallery from both seen and unseen categories
($\mathcal{P}^s \cup \mathcal{P}^u$),
introducing a \emph{seen-class bias} that makes retrieval harder.
We evaluate our method on both settings to demonstrate its generalization capability.

\subsection{Overview}
\begin{figure*}[t]
    \centering
    \resizebox{\textwidth}{!}{%
    \begin{minipage}{\textwidth}
        \centering
        \begin{subfigure}[b]{0.65\linewidth}
            \centering
            \includegraphics[width=\linewidth]{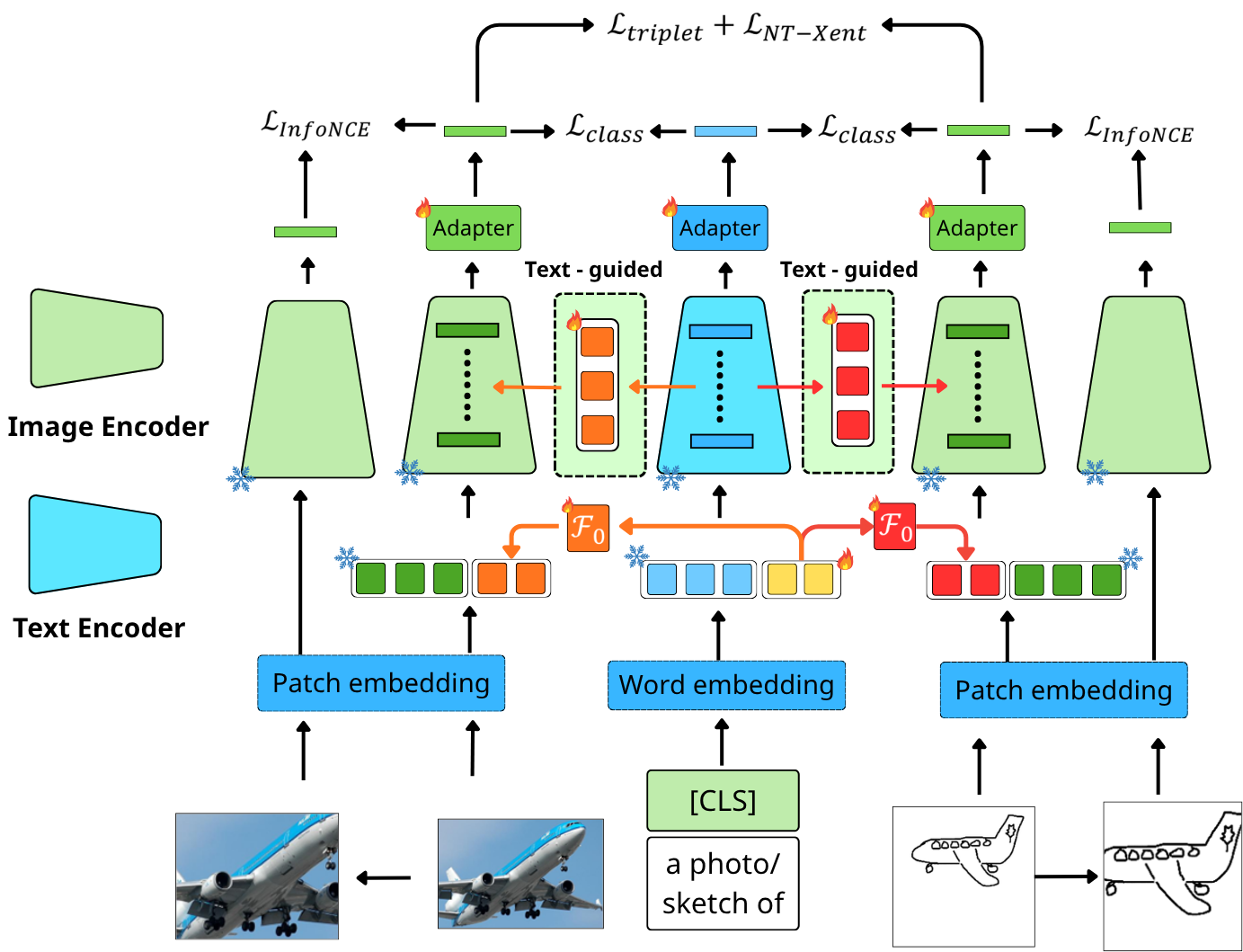}
            \caption{Overall framework}
            \label{fig:arch_a}
        \end{subfigure}
        \hfill
        \begin{subfigure}[b]{0.33\linewidth}
            \centering
            \includegraphics[width=\linewidth]{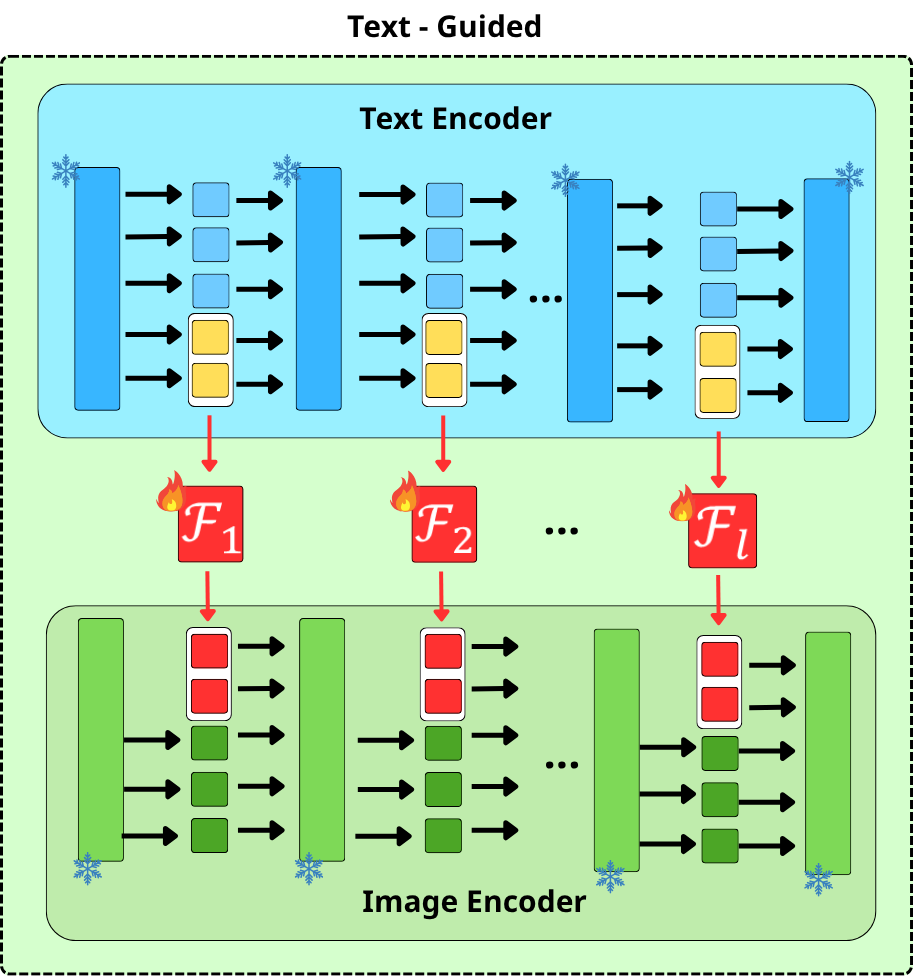}
            \caption{Text-guided coupling detail}
            \label{fig:arch_b}
        \end{subfigure}
    \end{minipage}%
    }
   \caption{Overview of SeCo-SBIR.
    (a)~The framework extends CLIP's dual-encoder architecture with three branches: a photo branch, a sketch branch, and a text branch, each followed by a lightweight adapter.
    Learnable prompt vectors are processed by the text encoder, and their intermediate representations are coupled into both visual encoders at every layer via the text-guided mechanism (dashed boxes).
    A frozen CLIP visual encoder (left, cyan) receives augmented inputs and provides the reference signal for the perturbation-based consistency constraint ($\mathcal{L}_{\mathrm{InfoNCE}}$).
    Cross-modal alignment is enforced through $\mathcal{L}_{\mathrm{triplet}}$, $\mathcal{L}_{\mathrm{NT\text{-}Xent}}$, and $\mathcal{L}_{\mathrm{class}}$.
    (b)~Detail of the text-guided coupling: at each transformer layer~$l$, the text encoder's intermediate prompt representations are projected into the visual encoder's embedding space via a learnable coupling function $\mathcal{F}_l$, producing layer-specific, semantically grounded visual prompts.
    Snowflake symbols denote frozen parameters.}
    \label{fig:arch}
\end{figure*}
 
We propose SeCo-SBIR, a semantically consistent prompt learning framework
that adapts CLIP~\cite{radford2021learning} for zero-shot sketch-based image retrieval.
Fig.~\ref{fig:arch} illustrates the overall architecture,
which extends CLIP's dual-encoder design with three parallel branches:
a photo branch, a sketch branch, and a text branch.
 
The framework addresses both sides of the adaptation - generalization tension through:
(i)~\textbf{text-guided multi-modal prompting} (\S\ref{sec:prompting}),
which routes learnable prompt vectors through the text encoder
and couples their intermediate representations into the visual encoder
at every layer via a learnable projection-injecting the text encoder's inherently transferable category-level semantics directly into the visual pathway;
(ii)~\textbf{lightweight adapters} (\S\ref{sec:adapters})
appended to each encoder branch;
(iii)~a \textbf{perturbation-based consistency constraint} (\S\ref{sec:consistency})
that aligns the adapted embeddings with a frozen CLIP reference via an InfoNCE loss over augmented and clean input pairs, counteracting the residual overfitting risk from the learnable components;
and (iv)~\textbf{cross-modal alignment losses} (\S\ref{sec:losses})
combining a triplet loss, an NT-Xent loss, and a classification loss.

\subsection{Text-Guided Multi-Modal Prompting}
\label{sec:prompting}

Standard visual prompt tuning~\cite{jia2022vpt} learns prompt tokens purely
within the visual encoder, without any explicit semantic grounding from language.
We instead adopt a text-guided approach where learnable prompts
first enter the text encoder and are then coupled into the visual encoder
at every layer.
This allows the visual prompts to inherit the text encoder's rich understanding
of object categories and cross-modal correspondences.
We apply this mechanism separately for the photo and sketch branches;
below, we describe the photo branch in detail-the sketch branch follows
identically with its own set of learnable prompts.

\subsubsection{Learnable Prompt Construction.}
We start from a fixed textual template such as \texttt{"a photo of a"},
whose word embeddings we denote
$W_0 = [w^1_0, w^2_0, \ldots, w^m_0] \in \mathbb{R}^{m \times d_t}$.
In standard CLIP, these fixed tokens are concatenated with the class-name
embedding $e_k \in \mathbb{R}^{d_t}$ and fed directly to the text encoder.
To adapt the textual pathway for the SBIR task, we introduce $b$ learnable
prompt vectors
$\mathbf{P}_0^t = \{p_0^{1,t}, p_0^{2,t}, \ldots, p_0^{b,t}\}$ with $p_0^{i,t} \in \mathbb{R}^{d_t}$,
where $b \leq m$.
These learnable vectors replace the first $b$ fixed word
embeddings in the template, while the remaining $m - b$ tokens are kept
intact.
The resulting prompt input to the text encoder for class~$k$ is:
\begin{equation}\label{eq:prompt}
  t_k
  = \bigl[\,
      p_0^{1,t},\;
      p_0^{2,t},\;
      \ldots,\;
      p_0^{b,t},\;
      w^{b+1}_0,\;
      \ldots,\;
      w^m_0,\;
      e_k
    \,\bigr].
\end{equation}
By retaining the latter portion of the original template, we preserve part
of the pre-trained semantic context while giving the model freedom to learn
task-specific representations in the first $b$ positions.
We share the same set of learnable vectors across all categories;
only $e_k$ varies per class.

\subsubsection{Cross-Modal Prompt Coupling.}
Rather than learning independent visual prompts,
we derive them from the text encoder's intermediate representations.
As the text encoder processes $t_k$ (Eq.~\ref{eq:prompt}) through its $L$ layers,
it produces updated prompt representations
$P^t_l \in \mathbb{R}^{b \times d_t}$ at each layer~$l$.
We project these into the visual embedding space via a learnable coupling function
$\mathcal{F}_l \colon \mathbb{R}^{d_t} \to \mathbb{R}^{d_v}$ (a linear projection jointly optimized with the other trainable parameters).
Since the photo and sketch branches maintain separate text-side prompts,
the text encoder produces branch-specific representations $P^{t,p}_l$ and $P^{t,s}_l$, yielding:
\begin{equation}\label{eq:coupling}
  P^p_l = \mathcal{F}_l(P^{t,p}_l), \qquad
  P^s_l = \mathcal{F}_l(P^{t,s}_l), \qquad
  l \in \{0, \ldots, L\}.
\end{equation}
Making $\mathcal{F}_l$ learnable rather than fixed allows the model to adapt the cross-modal mapping to the sketch--photo retrieval task, where the domain gap demands more than a faithful copy of the pre-trained text structure.

\subsubsection{Deep Prompt Injection into the Visual Encoder.}
At each visual encoder layer~$l$, we concatenate the coupled visual prompts
$P^p_l \in \mathbb{R}^{b \times d_v}$
(or $P^s_l$ for the sketch branch) with the existing patch tokens:
\begin{equation}\label{eq:inject}
  H_l
  = \mathrm{VisualLayer}_l
    \!\bigl(\,
      [c^v_{l-1},\, E_{l-1},\, P^p_l]
    \,\bigr),
\end{equation}
where $[c^v_{l-1}, E_{l-1}]$ denotes the \texttt{[CLS]} and patch tokens
from the previous layer.
After each layer, we discard the prompt token positions from~$H_l$
before feeding into layer~$l+1$,
preserving the visual encoder's original sequence length.
Fresh coupled prompts~$P^p_{l+1}$ are injected at every subsequent layer,
enabling progressive, semantically guided adaptation across the full depth
of the visual encoder.

We denote the prompted visual encoders as
$\hat{\mathcal{V}}_p(\cdot)$ and $\hat{\mathcal{V}}_s(\cdot)$
for photos and sketches, respectively.

\paragraph{Why text-guided prompting?}
CLIP's text encoder has already learned robust category-level semantics from large-scale language supervision - it can meaningfully represent categories never seen during SBIR training.
By deriving visual prompts from this encoder's intermediate representations, we inject inherently transferable semantic knowledge into the visual pathway at every layer, producing prompts that adapt the model to the sketch--photo domain while carrying category-level understanding that generalizes to unseen classes.
This contrasts with standard visual prompting~\cite{jia2022vpt}, where prompts learn solely from the visual training signal and are therefore tightly coupled to seen categories.

\subsection{Lightweight Adapter Modules}
\label{sec:adapters}
 
We attach a lightweight adapter to each encoder branch - $\mathcal{A}_p$, $\mathcal{A}_s$, and $\mathcal{A}_t$ for the photo, sketch, and text branches, respectively - to project features into a task-adapted embedding space.
Each adapter is a two-layer network:
\begin{equation}\label{eq:adapter}
  \mathcal{A}(\mathbf{z})
  = \Theta_2 \cdot \mathrm{ReLU}(\Theta_1 \cdot \mathbf{z}),
\end{equation}
where $\Theta_1 \in \mathbb{R}^{r \times d}$ and $\Theta_2 \in \mathbb{R}^{d' \times r}$.
The final adapted features are:
\begin{align}
  \hat{f}_p &= \mathcal{A}_p\!\bigl(\hat{\mathcal{V}}_p(p)\bigr), \label{eq:feat-p} \\
  \hat{f}_s &= \mathcal{A}_s\!\bigl(\hat{\mathcal{V}}_s(s)\bigr), \label{eq:feat-s} \\
  \hat{f}_t &= \mathcal{A}_t\!\bigl(\mathcal{T}(t_k)\bigr),          \label{eq:feat-t}
\end{align}
where $p$, $s$, and $t_k$ denote a photo, a sketch, and the prompt-augmented
text input for category~$k$, respectively.
 
\paragraph{Trainable components.}
Both CLIP encoders remain frozen throughout training. Adaptation is achieved entirely through the lightweight trainable modules:
(i)~the learnable prompt vectors $\mathbf{P}^{t}_0$;
(ii)~the coupling functions $\{\mathcal{F}_l\}_{l=0}^{L}$,
which project those representations into visual prompt space;
(iii)~the adapter modules $\mathcal{A}_p$, $\mathcal{A}_s$,
and $\mathcal{A}_t$;
and (iv)~the LayerNorm parameters $(\theta, \phi)$ within the encoder
transformer layers.

\subsection{Perturbation-Based Consistency Constraint}
\label{sec:consistency}

\begin{table*}[t]
\centering
\caption{Comparison for categorical ZS-SBIR across three benchmark datasets.
Best results are in \textbf{bold}, second-best are \underline{underlined}.}
\label{tab:zs_sbir}
\begin{tabular}{cl|cc|cc|cc|cc}
\hline
\multicolumn{2}{c|}{Methods} &
\multicolumn{2}{c|}{Sketchy-Ext-1} & \multicolumn{2}{c|}{Sketchy-Ext-2} & \multicolumn{2}{c|}{TU-Berlin-Ext} & \multicolumn{2}{c}{QuickDraw-Ext} \\
& & mAP@all & P@100 & mAP@200 & P@200 & mAP@all & P@100 & mAP@all & P@200 \\
\hline
\multirow{4}{*}{\rotatebox{90}{ViT}}
 & TVT~\cite{tian2022tvt} & 64.8 & 79.6 & 53.1 & 61.8 & 48.4 & 66.2 & 14.9 & 29.3 \\
 & PSKD~\cite{wang2022prototype}  & 68.8 & 78.6 & 56.0 & 64.5 & 50.2 & 66.2 & 15.0 & 29.8 \\
 & ZSE-RN~\cite{lin2023zeroshot}  & 69.8 & 79.7 & 52.5 & 62.4 & 54.2 & 65.7 & 14.5 & 21.6 \\
 & ZSE-Ret~\cite{lin2023zeroshot} & 73.6 & 80.8 & 50.4 & 60.2 & 56.9 & 63.7 & 14.2 & 20.2 \\
\hline
\multirow{7}{*}{\rotatebox{90}{CLIP}}
 & CLIP-AT~\cite{sain2023clip} & -- & -- & 72.3 & 72.5 & 65.1 & 73.2 & 20.2 & 38.8 \\
 & TLT~\cite{zhang2023task} & 77.9 & 84.3 & 66.1 & 73.0 & 61.5 & 69.5 & 27.8 & -- \\
 & Dr.CLIP~\cite{li2024drclip} & -- & -- & 74.0 & 70.6 & 68.5 & 76.3 & 24.2 & 31.2 \\
 & MARL~\cite{lyou2024modality} & -- & -- & 69.1 & 75.5 & 70.5 & 77.7 & 32.7 & 42.5 \\
 & MCG~\cite{zhou2024mcg} & 78.4 & 84.2 & \underline{79.0} & 75.7 & 64.2 & 75.0 & 31.4 & 40.2 \\
 & SpLIP~\cite{singha2024splip} & \underline{80.2} & \underline{86.7} & 76.4 & \underline{77.3} & \underline{73.1} & \underline{78.2} & \underline{34.2} & \underline{44.6} \\
& DCDL~\cite{li2025dcdl} & -- & -- & 72.6 & 76.9 & 63.4 & 74.1 & 33.6 & 29.6 \\
& SketchFusion~\cite{koley2025sketchfusion}  & -- & -- & 76.1 & 76.3 & 69.5 & 75.3 & 24.2 & 39.9 \\
\hline
 & \textbf{SeCo-SBIR} & \textbf{80.6} & \textbf{87.0} & \textbf{80.0} & \textbf{77.5} & \textbf{78.7} & \textbf{84.7} & \textbf{36.0} & \textbf{46.1} \\
\hline
\end{tabular}%
\end{table*}

\begin{table}[t]
\centering
\caption{Comparison for GZS-SBIR.}
\label{tab:gzs_sbir}
\resizebox{\columnwidth}{!}{%
\begin{tabular}{l|cc|cc}
\hline
\multirow{2}{*}{Method} & \multicolumn{2}{c|}{Sketchy-Ext-2} & \multicolumn{2}{c}{TU-Berlin-Ext} \\
 & mAP@200 & P@200 & mAP@all & P@100 \\
\hline
SEM-PCYC~\cite{dutta2019semantically} & -- & -- & 19.2 & 29.8 \\
OCEAN~\cite{zhu2020ocean} & -- & -- & 31.2 & 34.1 \\
ZSE-RN~\cite{lin2023zeroshot} & -- & -- & 43.2 & 46.0 \\
ZSE-Ret~\cite{lin2023zeroshot} & -- & -- & 46.4 & 48.5 \\
STL~\cite{ge2023semitransductive} & 63.4 & 53.8 & 40.2 & 49.8 \\
CLIP-AT~\cite{sain2023clip} & 55.6 & 62.7 & 60.9 & 63.8 \\
MARL~\cite{lyou2024modality} & 62.3 & 68.5 & 62.6 & 67.8 \\
Dr.CLIP~\cite{li2024drclip} & \underline{73.9} & 71.4 & 65.3 & 68.5 \\
SpLIP~\cite{singha2024splip} & 68.2 & \underline{74.5} & \underline{66.7} & \underline{70.3} \\
\hline
\textbf{SeCo-SBIR} & \textbf{77.6} & \textbf{75.1} & \textbf{71.3} & \textbf{77.8} \\
\hline
\end{tabular}%
}
\end{table}

\begin{table}[t]
\centering
\caption{Across-dataset ZS-SBIR. The model is trained on Sketchy-Ext
and tested on TU-Berlin-Ext and QuickDraw-Ext.}
\label{tab:across_dataset}
\resizebox{\columnwidth}{!}{%
\begin{tabular}{l|cc|cc}
\hline
\multirow{2}{*}{Method} & \multicolumn{2}{c|}{TU-Berlin-Ext} & \multicolumn{2}{c}{QuickDraw-Ext} \\
 & mAP@all & P@100 & mAP@all & P@100 \\
\hline
CC-DG~\cite{shankar2018generalizing} & 30.8 & 43.4 & 15.6 & 22.7 \\
SAKE~\cite{liu2019semantic} & 38.9 & 50.6 & 17.4 & 24.2 \\
ZSE-RN~\cite{lin2023zeroshot} & 47.6 & 59.0 & 22.8 & 33.8 \\
CLIP-AT~\cite{sain2023clip} & 56.4 & 63.1 & 30.7 & 45.0 \\
SpLIP~\cite{singha2024splip} & \underline{70.6} & \underline{76.0} & \underline{45.8} & \underline{58.6} \\
\hline
\textbf{SeCo-SBIR} & \textbf{75.6} & \textbf{79.5} & \textbf{52.6} & \textbf{60.0} \\
\hline
\end{tabular}%
}
\end{table}

To regularize the trainable model and preserve CLIP's zero-shot transfer
capability, we introduce a consistency constraint that aligns the adapted
embeddings with those of a fully frozen CLIP visual encoder~$\mathcal{V}_{\mathrm{frz}}$
(without any prompts or adapters).

\paragraph{Dual-branch design with input perturbation.}
For a given photo~$p$, we generate a perturbed view
$\tilde{p} = \mathcal{T}_{\mathrm{aug}}(p)$
via stochastic data augmentation (random erase, center crop,
and horizontal flipping).
The augmented input feeds the frozen reference encoder,
while the clean input feeds the trainable (prompted + adapted) encoder:
\begin{equation}\label{eq:dual-branch}
  f^{\mathrm{frz}}_p = \mathcal{V}_{\mathrm{frz}}(\tilde{p}),
  \qquad
  \hat{f}_p = \mathcal{A}_p\!\bigl(\hat{\mathcal{V}}_p(p)\bigr).
\end{equation}

This asymmetric design serves two purposes.
First, it prevents the trivial solution that would arise if both branches
received the same input:
the trainable model could minimize the consistency loss simply by replicating
the frozen model's behavior, learning nothing task-specific.
The augmentation breaks this degeneracy by ensuring that the two branches
see different views of the same sample.
Second, by feeding augmented inputs to the frozen branch,
we create a form of soft supervision:
the trainable model must produce clean-input embeddings that are structurally
compatible with the frozen model's representations under input variations,
encouraging robustness without constraining the model to an exact copy
of the frozen output.

\paragraph{Consistency via InfoNCE.}
We align the two branches using an InfoNCE loss~\cite{oord2018infonce, wang2020understanding, parulekar2023infonce, parulekar2023infonce, awasthi2022negative} rather than pointwise alternatives (MSE, L1).
InfoNCE treats alignment as a discrimination problem - each adapted embedding must be closer to its corresponding frozen embedding than to those of other samples - preserving the relative structure of CLIP's embedding space, which matters more for retrieval than absolute distances.
For the photo branch over a mini-batch of $B$ samples:
\begin{equation}\label{eq:infonce-p}
  \mathcal{L}^{p}_{\mathrm{InfoNCE}}
  = -\frac{1}{B}\sum_{i=1}^{B}
    \log \frac{
      \exp\!\bigl(\mathrm{sim}(f^{\mathrm{frz}}_{p_i},\,\hat{f}_{p_i})/\tau_c\bigr)
    }{
      \sum_{j=1}^{B}
      \exp\!\bigl(\mathrm{sim}(f^{\mathrm{frz}}_{p_i},\,\hat{f}_{p_j})/\tau_c\bigr)
    },
\end{equation}
where $\tau_c$ is a temperature parameter.
An analogous loss $\mathcal{L}^{s}_{\mathrm{InfoNCE}}$ is defined for the
sketch branch, with
$f^{\mathrm{frz}}_{s_i} = \mathcal{V}_{\mathrm{frz}}(\tilde{s}_i)$
and
$\hat{f}_{s_i} = \mathcal{A}_s(\hat{\mathcal{V}}_s(s_i))$.
The total consistency loss combines both:
\begin{equation}\label{eq:consist}
  \mathcal{L}_{\mathrm{consist}}
  = \mathcal{L}^{p}_{\mathrm{InfoNCE}}
  + \mathcal{L}^{s}_{\mathrm{InfoNCE}}.
\end{equation}

\subsection{Cross-Modal Alignment Losses}
\label{sec:losses}

In addition to the consistency constraint, we employ three complementary
losses that build a discriminative embedding space bridging the sketch-photo
modality gap.

\subsubsection{Cross-Modal Triplet Loss.}
The triplet loss~\cite{schroff2015facenet, yu2018correcting, ge2018deep, do2019theoretically} pulls the adapted sketch and photo
embeddings of the same category together while pushing apart those from
different categories.
Given an anchor sketch~$\hat{f}_s$, a positive photo~$\hat{f}^+_p$
(same category), and a negative photo~$\hat{f}^-_p$ (different category):
\begin{equation}\label{eq:triplet}
  \mathcal{L}_{\mathrm{triplet}}
  = \frac{1}{B}\sum_{i=1}^{B}
    \max\!\bigl(0,\;
      \mu + d(\hat{f}_{s_i},\,\hat{f}^+_{p_i})
           - d(\hat{f}_{s_i},\,\hat{f}^-_{p_i})
    \bigr),
\end{equation}
where
$d(\mathbf{a},\mathbf{b}) = 1 - \frac{\mathbf{a} \cdot \mathbf{b}}
                                       {\|\mathbf{a}\|\,\|\mathbf{b}\|}$
is the cosine distance and $\mu > 0$ is the margin.

\subsubsection{Normalized Temperature-Scaled Cross-Entropy (NT-Xent).}
The NT-Xent loss~\cite{chen2020simclr, wang2021understanding} provides a batch-level contrastive
signal that complements the triplet loss by treating all non-matching pairs
in the batch as negatives.
We define the sketch-to-photo and photo-to-sketch directions as:
\begin{align}
  \ell^{(i)}_{s \to p}
  &= -\log \frac{
      \exp\!\bigl(\mathrm{sim}(\hat{f}_{s_i},\,\hat{f}_{p_i})/\tau\bigr)
    }{
      \sum_{j=1}^{B}
      \exp\!\bigl(\mathrm{sim}(\hat{f}_{s_i},\,\hat{f}_{p_j})/\tau\bigr)
    }, \label{eq:ntxent-sp} \\[4pt]
  \ell^{(i)}_{p \to s}
  &= -\log \frac{
      \exp\!\bigl(\mathrm{sim}(\hat{f}_{p_i},\,\hat{f}_{s_i})/\tau\bigr)
    }{
      \sum_{j=1}^{B}
      \exp\!\bigl(\mathrm{sim}(\hat{f}_{p_i},\,\hat{f}_{s_j})/\tau\bigr)
    }, \label{eq:ntxent-ps}
\end{align}
where $\tau$ is the temperature.
The full NT-Xent loss averages both directions:
\begin{equation}\label{eq:ntxent}
  \mathcal{L}_{\mathrm{NT\text{-}Xent}}
  = \frac{1}{2B}\sum_{i=1}^{B}
    \bigl(\ell^{(i)}_{s \to p} + \ell^{(i)}_{p \to s}\bigr).
\end{equation}
This bidirectional formulation ensures that each sketch is close to its
matching photo and each photo is close to its matching sketch,
offering a richer supervisory signal than the triplet loss alone.

\subsubsection{Classification Loss.}
We construct class prototypes by passing the prompt-augmented text input
$t_k$ (Eq.~\ref{eq:prompt}) for each seen category
$c_k \in \mathcal{C}^s$ through the text encoder and its adapter,
obtaining $\{\hat{f}^k_t\}_{k=1}^{|\mathcal{C}^s|}$.
The classification loss for modality $\mathtt{I} \in \{p, s\}$ is:
\begin{equation}\label{eq:cls}
  \mathcal{L}^{\mathtt{I}}_{\mathrm{class}}
  = -\frac{1}{B}\sum_{i=1}^{B}
    \log \frac{
      \exp\!\bigl(\mathrm{sim}(\hat{f}^{\mathtt{I}}_i,\,
                                \hat{f}^{y_i}_t) / \tau_{\mathrm{cls}}\bigr)
    }{
      \sum_{k=1}^{|\mathcal{C}^s|}
      \exp\!\bigl(\mathrm{sim}(\hat{f}^{\mathtt{I}}_i,\,
                                \hat{f}^{k}_t) / \tau_{\mathrm{cls}}\bigr)
    },
\end{equation}
where $y_i$ is the ground-truth label,
$\mathrm{sim}(\cdot,\cdot)$ denotes cosine similarity,
and $\tau_{\mathrm{cls}}$ is the temperature.
Because the class prototypes are produced by the same learnable prompts as the visual features, they co-adapt during training; at test time, we compose prototypes for unseen category names using the same prompt structure, enabling zero-shot transfer.
The total classification loss is
$\mathcal{L}_{\mathrm{class}} = \mathcal{L}^{p}_{\mathrm{class}} + \mathcal{L}^{s}_{\mathrm{class}}$.
\subsection{Overall Training Objective}

The full training objective combines all losses:
\begin{equation}\label{eq:total}
  \mathcal{L}_{\mathrm{total}}
  = \mathcal{L}_{\mathrm{triplet}}
  + \mathcal{L}_{\mathrm{NT\text{-}Xent}}
  + \lambda_1 \mathcal{L}_{\mathrm{class}}
  + \lambda_2 \mathcal{L}_{\mathrm{consist}},
\end{equation}
where $\lambda_1$ and $\lambda_2$ control the relative importance of the
classification and consistency terms.

\paragraph{Inference.}
At test time, we discard the augmentation pipeline and the frozen reference branch.
For a query sketch $s_u \in \mathcal{S}^u$ from an unseen category,
we extract the adapted feature
$\hat{f}_{s_u} = \mathcal{A}_s(\hat{\mathcal{V}}_s(s_u))$
and rank gallery photos by cosine similarity.
For ZS-SBIR the gallery consists of $\mathcal{P}^u$;
for GZS-SBIR it consists of $\mathcal{P}^s \cup \mathcal{P}^u$.

\begin{table}[t]
\centering
\caption{Ablation of prompt and text-guided design.}
\label{tab:ablation_prompt}
\resizebox{\columnwidth}{!}{
\begin{tabular}{l|cc}
\hline
 & mAP@all & P@100 \\
\hline
CLIP (w/o prompt) & 72.4 & 79.2 \\
CLIP + Text Prompt only & 73.9 & 80.7 \\
CLIP + Image Prompt only & 74.8 & 81.3 \\
CLIP + Text/Image Prompt (w/o text-guided) & 76.1 & 82.3 \\
CLIP + Text/Image Prompt (w/ text-guided) & 77.5 & 83.5 \\
CLIP + Text/Image Prompt (w/ text-guided + adapter) & \textbf{78.7} & \textbf{84.7} \\
\hline
\end{tabular}
}
\end{table}

\begin{table}[t]
\centering
\caption{Ablation of loss components.}
\label{tab:ablation_loss}
\resizebox{\columnwidth}{!}{%
\begin{tabular}{cccc|cc}
\hline
$\mathcal{L}_{\mathrm{triplet}}$ & $\mathcal{L}_{\mathrm{consist}}$ & $\mathcal{L}_{\mathrm{class}}$ & $\mathcal{L}_{\mathrm{NT\text{-}Xent}}$ & mAP@all & P@100 \\
\hline
\checkmark & \texttimes & \texttimes & \texttimes & 62.2 & 73.3 \\
\checkmark & \checkmark & \texttimes & \texttimes & 67.3 & 78.2 \\
\checkmark & \checkmark & \checkmark & \texttimes & 72.8 & 81.9 \\
\checkmark & \checkmark & \checkmark & \checkmark & \textbf{78.7} & \textbf{84.7} \\
\hline
\end{tabular}%
}
\end{table}

\begin{table}[t]
\centering
\caption{Ablation of consistency constraint and perturbation.}
\label{tab:ablation_consistency_perturbation}
\begin{tabular}{l|cc}
\hline
Configuration & mAP@all & P@100 \\
\hline
w/o consistency & 76.1 & 81.8 \\
w/o perturbation & 77.2 & 83.2 \\
\textbf{Our} & \textbf{78.7} & \textbf{84.7} \\
\hline
\end{tabular}
\end{table}

\begin{table}[t]
\centering
\caption{Effect of consistency loss function.}
\label{tab:ablation_criterion}
\begin{tabular}{l|cc}
\hline
 & mAP@all & P@100 \\
\hline
MSE & 77.3 & 83.5 \\
Cosine & 76.4 & 82.0 \\
L1 & 76.9 & 82.7 \\
InfoNCE & \textbf{78.7} & \textbf{84.7} \\
\hline
\end{tabular}
\end{table}

\section{Experiments and Results}

\textbf{Datasets.}
We evaluate our method on three publicly available benchmarks for SBIR:
Sketchy-Ext~\cite{sangkloy2016sketchy, dutta2019semantically, yelamarthi2018zero},
TU-Berlin-Ext~\cite{eitz2012humans, zhang2016sketchnet, liu2017deep},
and QuickDraw-Ext~\cite{ha2018neural, dey2019doodle}.
Sketchy-Ext enlarges the original Sketchy~\cite{shankar2018generalizing},
which includes more than 73,000 images across 125 classes.
Following~\cite{sain2023clip}, 21 categories absent from ImageNet~\cite{imagenet}
are reserved for testing, and the remaining 104 categories are used for training.
TU-Berlin-Ext incorporates more than 204,000 natural images spanning 250 categories.
Following~\cite{sain2023clip}, we split 220 classes for training and 30 for testing,
ensuring that each class has at least 400 images.
QuickDraw-Ext is a large-scale dataset for ZS-SBIR containing 110 categories
with 330,000 sketches (3,000 per category) and 204,000 Flickr-sourced images.
We reserve 30 test classes not present in ImageNet and use 80 classes for training,
following the protocol of~\cite{dey2019doodle}.

\textbf{Implementation Details.}
We implement the model in PyTorch on a single NVIDIA H100 GPU.
The CLIP model~\cite{radford2021learning} with ViT-B/32 weights serves as
the backbone for both sketch and photo encoders.
Input images are resized to $224 \times 224$.
The margin parameter is set to $\mu = 0.3$.
We train for 10 epochs using SGD with momentum,
a learning rate of $2 \times 10^{-5}$, and a batch size of 32. We use the single-sentence template following~\cite{singha2024splip}
to construct the textual prompts for our method. Values of $\lambda_{1, 2}$ are set to 1 and 1, respectively

\textbf{Evaluation Metrics.}
Following common practice in SBIR evaluation~\cite{zhou2024mcg, singha2024splip},
we assess retrieval performance over the top 200 returned images.
We report mean Average Precision (mAP@all) together with Precision at 200 (P@200).
In addition, to remain consistent with recent work,
we use P@100 for TU-Berlin-Ext and Sketchy-Ext-1,
and mAP@200 for Sketchy-Ext-2.
These metrics provide a reliable and standard way to measure retrieval quality
across datasets.

\subsection{Results and Analysis}
\label{sec:results}

\subsubsection{Comparison with State-of-the-Art}
\label{sec:sota}

We compare \textbf{SeCo-SBIR} against a comprehensive set of state-of-the-art
methods spanning two backbone families:
(i)~ViT-based approaches and (ii)~CLIP-based approaches.

\noindent\textbf{Categorical ZS-SBIR.}
Table~\ref{tab:zs_sbir} presents the categorical ZS-SBIR results on two splits of Sketchy-Ext (Sketchy-Ext-1 and Sketchy-Ext-2), TU-Berlin-Ext, and QuickDraw-Ext.
SeCo-SBIR achieves the highest scores on every metric across all four benchmarks,
obtaining 80.6\%/87.0\% on Sketchy-Ext-1, 80.0\%/77.5\% on Sketchy-Ext-2,
78.7\%/84.7\% on TU-Berlin-Ext, and 36.0\%/46.1\% on QuickDraw-Ext.
Compared with the strongest prior method SpLIP~\cite{singha2024splip},
SeCo-SBIR yields consistent improvements,
with especially clear gains on TU-Berlin-Ext ($+5.6$\% mAP@all, $+6.5$\% P@100)
and QuickDraw-Ext ($+1.8$\% mAP@all, $+1.5$\% P@200).
All CLIP-based methods substantially outperform ViT-based counterparts,
confirming the value of pre-trained vision-language representations for ZS-SBIR,
and SeCo-SBIR establishes a clear lead within the CLIP family.
The largest gains appear on TU-Berlin-Ext, which has 250 categories - the
most fine-grained label space among the benchmarks - making it particularly
sensitive to seen-class overfitting;
the consistency constraint in SeCo-SBIR anchors adapted representations to
CLIP's pre-trained space, preserving transferability to unseen classes.
On QuickDraw-Ext, where sketches are highly abstract and noisy,
SeCo-SBIR still ranks first, indicating that the text-guided prompting
mechanism provides robust semantic grounding even under degraded visual quality.

\noindent\textbf{Generalized ZS-SBIR.}
Table~\ref{tab:gzs_sbir} reports the GZS-SBIR results,
where the gallery comprises photos from both seen and unseen categories.
SeCo-SBIR achieves the best performance on both benchmarks,
reaching 77.6\% mAP@200 and 75.1\% P@200 on Sketchy-Ext-2,
and 71.3\% mAP@all and 77.8\% P@100 on TU-Berlin-Ext.
Notably, the improvement margins over SpLIP are even larger than in
the standard ZS-SBIR setting,
particularly on TU-Berlin-Ext ($+4.6$\% mAP@all, $+7.5$\% P@100),
indicating that the perturbation-based consistency constraint effectively
mitigates the seen-class bias inherent to GZS-SBIR by maintaining
a balanced representation that does not disproportionately favor seen categories.
\begin{figure*}[t]
\centering
\includegraphics[width=0.9\textwidth]{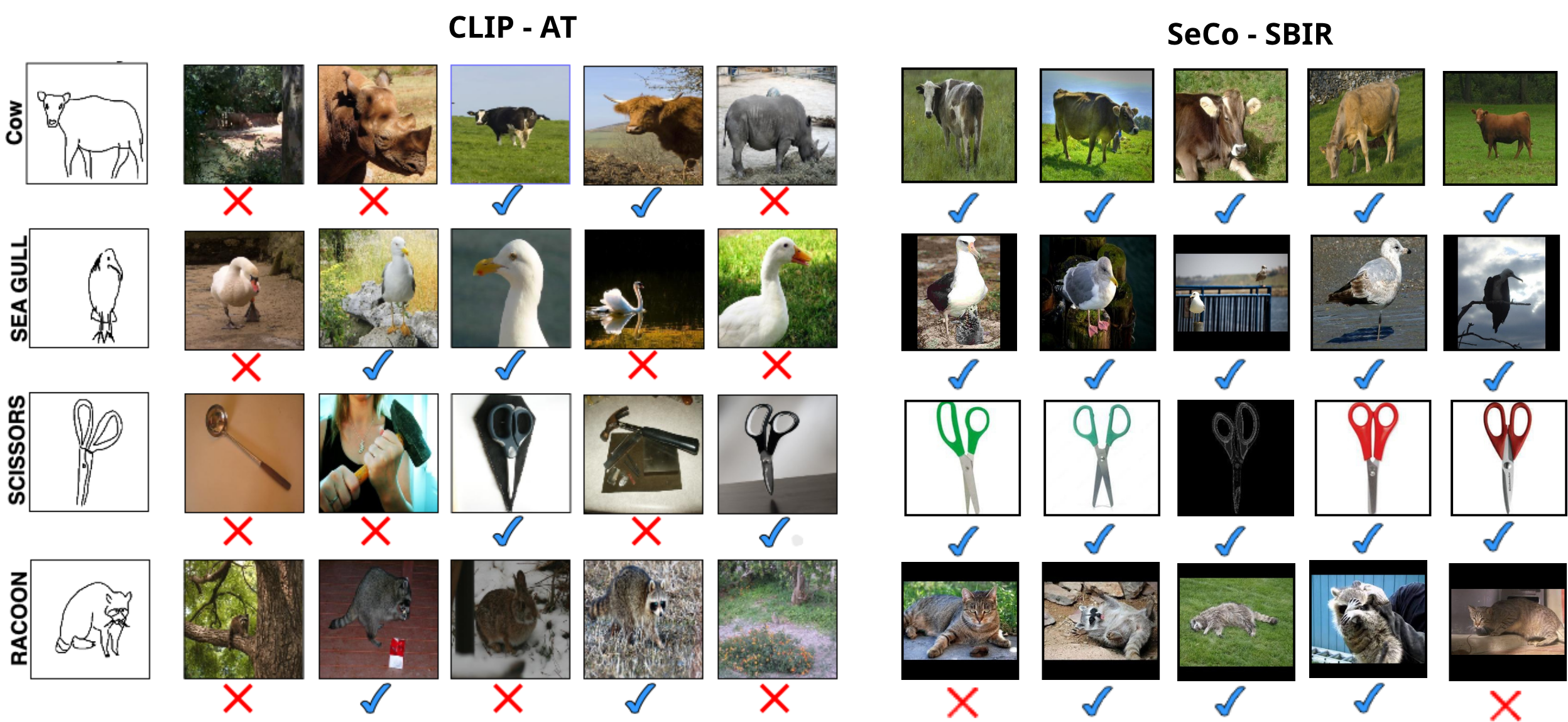}
\caption{Qualitative retrieval results on Sketchy-Ext-2.
For each query sketch (left), we show the top-5 retrieved photos
by CLIP-AT (middle) and our method (right).
Blue check marks indicate correct retrievals.}
\label{fig:qualitative_retrieval}
\end{figure*}

\noindent\textbf{Across-Dataset ZS-SBIR.}
Following~\cite{lin2023zeroshot}, we train on Sketchy-Ext and evaluate directly
on the unseen test classes of TU-Berlin-Ext and QuickDraw-Ext without fine-tuning
(Table~\ref{tab:across_dataset}).
SeCo-SBIR ranks first on both targets, achieving 75.6/79.5 on TU-Berlin-Ext
and 52.6/60.0 on QuickDraw-Ext, surpassing SpLIP by $+5.0\%/+3.5\%$ and
$+6.8\%/+1.4\%$, respectively.
This setting tests robustness to both category and dataset shift simultaneously;
the large $+6.8\%$ mAP@all gain on QuickDraw-Ext - where the domain gap is
largest - confirms that the consistency constraint preserves CLIP's
distributional robustness, while the text-guided prompting provides
stable semantic grounding that transfers across visual domains.
\subsubsection{Ablation Analysis}
\label{sec:ablation}

We conduct comprehensive ablation studies on the TU-Berlin dataset
to isolate the contribution of each proposed component.

\vspace{0.5em}
\noindent\textbf{Effect of prompt and text-guided design.}
Table~\ref{tab:ablation_prompt} shows the contribution of each design component
on TU-Berlin-Ext. Starting from the frozen CLIP baseline (72.4\% mAP@all),
combining independent text and image prompts raises performance to 76.1\% ($+3.7\%$).
The text-guided coupling mechanism adds a further $+1.4\%$,
confirming that routing textual semantics into the visual encoder via
$\mathcal{F}_l$ provides complementary information beyond what independent
prompting achieves.
Appending the lightweight adapters contributes another $+1.2\%$,
bringing the full model to 78.7\% mAP@all.

\vspace{0.5em}
\noindent\textbf{Effect of individual loss components.}
Table~\ref{tab:ablation_loss} shows the impact of progressively adding
each loss term. Training with $\mathcal{L}_{\mathrm{triplet}}$ alone yields 62.2\% mAP@all.
Adding $\mathcal{L}_{\mathrm{consist}}$ produces $+5.1\%$, demonstrating
the importance of anchoring adapted representations to the frozen CLIP space.
$\mathcal{L}_{\mathrm{class}}$ further contributes $+5.5\%$ through
semantic transfer from the co-adapted text prototypes,
and $\mathcal{L}_{\mathrm{NT\text{-}Xent}}$ adds $+5.9\%$ via dense
batch-level contrastive supervision.
All four losses contribute substantially and non-redundantly.

\vspace{0.5em}
\noindent\textbf{Effect of consistency constraint and perturbation.}
Table~\ref{tab:ablation_consistency_perturbation} isolates the contributions
of the consistency constraint and the perturbation strategy. Removing the consistency constraint entirely causes a $-2.6\%$ mAP@all drop,
confirming its critical role in regularizing the adapted model against
overfitting to seen categories.
When consistency is applied without perturbation
(i.e., both branches receive clean inputs), performance recovers partially
(77.2\%) but remains $1.5$ points below the full model.
This validates the asymmetric perturbation design:
feeding augmented inputs to the frozen branch prevents the trivial solution
of copying the reference encoder and instead forces the trainable branch
to learn input-invariant representations,
yielding stronger generalization to unseen classes.

\vspace{0.5em}
\noindent\textbf{Choice of consistency loss function.}
Table~\ref{tab:ablation_criterion} compares several loss functions for
the consistency constraint. InfoNCE outperforms all pointwise alternatives (MSE, Cosine, L1)
by $+1.4\%$ to $+2.3\%$ mAP@all.
Unlike pointwise losses that penalize absolute embedding distances,
InfoNCE preserves the \emph{relative} structure of the embedding space
by framing alignment as a discrimination task - a formulation better
suited for retrieval, where ranking order matters more than
absolute embedding values.

\subsubsection{Qualitative Comparisons}
\label{sec:qualitative}
 
Fig.~\ref{fig:qualitative_retrieval} compares top-$k$ retrieval results of SeCo-SBIR and CLIP-AT~\cite{sain2023clip} on unseen categories.
SeCo-SBIR consistently retrieves correct photos in the top ranks, especially for visually ambiguous classes where sketch abstraction makes it hard to distinguish semantically similar categories---a benefit of grounding visual representations in category-level semantics via text-guided prompting.


\section{Conclusion}

We have presented SeCo-SBIR, a prompt learning framework for ZS-SBIR that addresses the tension between task adaptation and zero-shot generalization from both sides: text-guided multi-modal prompting adapts the visual encoder through the text encoder's inherently transferable category-level semantics, while a perturbation-based consistency constraint regularizes the remaining learnable capacity against seen-class drift.
Experiments on three standard benchmarks show state-of-the-art results across categorical, generalized, and across-dataset settings, with the largest gains where overfitting and domain shift are most severe.
Ablation studies confirm that each component contributes positively and non-redundantly.
More broadly, our results suggest that channeling adaptation through language-grounded knowledge while explicitly constraining learnable components is an effective principle for deploying vision--language models in settings that demand generalization to unseen categories.

\begin{acks}
This research is funded by the Posts and Telecommunications Institute of Technology (PTIT). The authors would like to thank PTIT for the financial support.
\end{acks}

{\small
\bibliographystyle{ACM-Reference-Format}
\bibliography{sample-base}
}

\clearpage
\appendix

\section*{Appendix}
In this supplementary document, we present detailed information and further experimental results.

\section{Additional ablation studies}
\noindent\textbf{Effect of the consistency constraint design.}
To validate the design choices behind the perturbation-based consistency
constraint, we compare several variants in Table~\ref{tab:ablation_consistency}.
\begin{table}[H]
\centering
\caption{Ablation of the consistency constraint design on Tu-Berlin-Ext.}
\label{tab:ablation_consistency}
\begin{tabular}{l|cc}
\hline
 & mAP@all & P@100 \\
\hline
 No consistency & 76.1 & 81.8 \\
 Photo consistency only & 77.2 & 82.8 \\
 Sketch consistency only & 77.5 & 83.4 \\
 Sketch and photo consistency & \textbf{78.7} & \textbf{84.7} \\
\hline
\end{tabular}
\end{table}
Removing the consistency constraint entirely (first row) causes
a $-2.6\%$ mAP@all drop, confirming its regularization benefit.
Applying consistency to only one modality recovers part of the gain:
the sketch-only variant ($+1.3\%$) slightly outperforms the photo-only variant ($+1.1\%$),
likely because sketch representations are more prone to distributional
drift due to the larger domain gap between sketches and CLIP's
pre-training distribution.
Applying consistency to both branches yields the best result,
indicating that regularizing each modality independently contributes
complementary benefits.

\vspace{0.5em}
\noindent\textbf{Effect of the Adapter module.} Table \ref{tab:ablation_adapter} shows that introducing adapters into both branches yields the best performance on TU-Berlin-Ext. Specifically, without any adapter, the model achieves 77.5\% mAP@all and 83.5\% P@100. Adding only the Text Adapter slightly improves the results to 77.9\% mAP@all and 84.0\% P@100. Similarly, using only the Image Adapter further enhances the performance, reaching 78.3\% mAP@all and 84.2\% P@100, which is higher than the Text-Adapter-only setting on both metrics. Notably, the configuration that employs both the Text Adapter and the Image Adapter achieves the best overall results, with 78.7\% mAP@all and 84.7\% P@100. Compared with the model without adapters, this setting brings an improvement of 1.2\% in mAP@all and 1.2\% in P@100. These results indicate that each adapter individually contributes positively to retrieval performance, while combining both provides the most substantial and consistent gain.

\begin{table}[H]
\centering
\caption{Ablation of the Adapter module on Tu-Berlin-Ext.}
\label{tab:ablation_adapter}
\begin{tabular}{l|cc}
\hline
 & mAP@all & P@100 \\
\hline
 W/o Adapter & 77.5 & 83.5 \\
 Text Adapter only & 77.9 & 84.0 \\
 Image Adapter only & 78.3 & 84.2 \\
 Both Adapter & \textbf{78.7} & \textbf{84.7} \\
\hline
\end{tabular}
\end{table}

\vspace{0.5em}
\noindent\textbf{The number of learnable tokens.} Table \ref{tab:ablation_number_tokens} shows that the model performs best when the number of learnable tokens is kept small. Specifically, the 1-token setting achieves the highest mAP@all of 79.2\%, while the 2-token setting gives the best P@100 of 84.7\%. Overall, the configurations with 1 to 3 tokens maintain strong performance, with mAP@all ranging from 78.7 to 79.2 and P@100 ranging from 84.1\% to 84.7\%. In contrast, when the number of tokens is further increased to 4, 5, and 6, the performance gradually declines on both metrics. The mAP@all drops to 78.6\%, 77.9\%, and 77.6\%, respectively, while P@100 decreases to 84.2\%, 83.8\%, and 83.2\%. This trend suggests that using too many learnable tokens does not provide further benefits for the model on TU-Berlin.

\begin{table}[H]
\centering
\caption{Ablation of the number of learnable tokens on Tu-Berlin-Ext.}
\label{tab:ablation_number_tokens}
\begin{tabular}{c|cc}
\hline
Number of tokens & mAP@all & P@100 \\
\hline
 1 & \textbf{79.2} & 84.1 \\
 2 & 78.7 & \textbf{84.7} \\
 3 & 78.9 & 84.5 \\
 4 & 78.6 & 84.2 \\
 5 & 77.9 & 83.8 \\
 6 & 77.6 & 83.2 \\
\hline
\end{tabular}
\end{table}

\vspace{0.5em}
\noindent\textbf{The impact of loss function weights.} Table \ref{tab:ablation_loss_weight} shows that the model performance is quite sensitive to the balance between the two loss weights, $\lambda_1$ and $\lambda_2$. When fixing $\lambda_1 = 1$ and gradually increasing $\lambda_2$ from 0 to 1, the results improve clearly compared with the setting that uses only one loss component, and the configuration $(1,1)$ achieves the best performance with 78.7\% mAP@all and 84.7\% P@100. This indicates that using both loss components together is more effective than using an unbalanced setting. In addition, when fixing $\lambda_2 = 1$ and changing $\lambda_1$ from 0 to 1, the performance also improves steadily, with mAP@all increasing from 77.8\% to 78.7\% and P@100 rising from 83.6\% to 84.7\%. Overall, the settings in which one weight is too small, such as $(1,0)$ or $(0,1)$, give noticeably lower results. This trend suggests that the two loss components play complementary roles, and the model performs best when both are assigned balanced weights.

\begin{table}[H]
\centering
\caption{The impact of loss weight on Tu-Berlin-Ext.}
\label{tab:ablation_loss_weight}
\begin{tabular}{cc|cc}
\hline
$\lambda_1$ & $\lambda_2$ & mAP@all & P@100 \\
\hline
 1 & 0 & 77.4 & 83.7 \\
 1 & 0.2 & 78.7 & 84.7 \\
 1 & 0.4 & 77.1 & 84.4 \\
 1 & 0.6 & 77.9 & 84.5 \\
 1 & 0.8 & 78.4 & 84.6 \\
 0 & 1 & 77.8 & 83.6 \\
 0.2 & 1 & 78.2 & 84.0 \\
 0.4 & 1 & 78.4 & 84.2 \\
 0.6 & 1 & 78.6 & 84.4 \\
 0.8 & 1 & 78.6 & 84.6 \\
 1 & 1 & \textbf{78.7} & \textbf{84.7} \\
\hline
\end{tabular}
\end{table}

\begin{figure*}
    \centering
    \includegraphics[width=\linewidth]{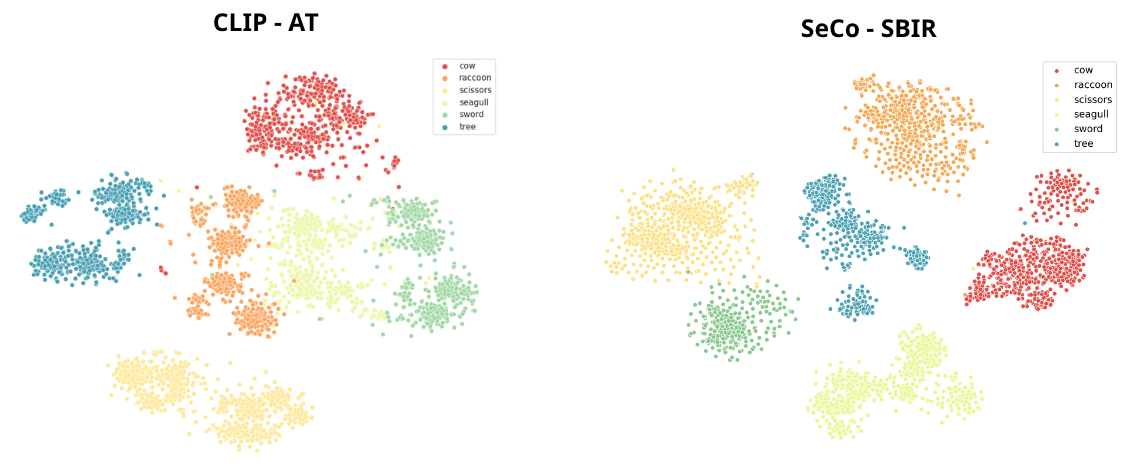}
    \caption{t-SNE visualization of photo domain}
    \label{fig:tsne_photo}
\end{figure*}

\begin{figure*}
    \centering
    \includegraphics[width=\linewidth]{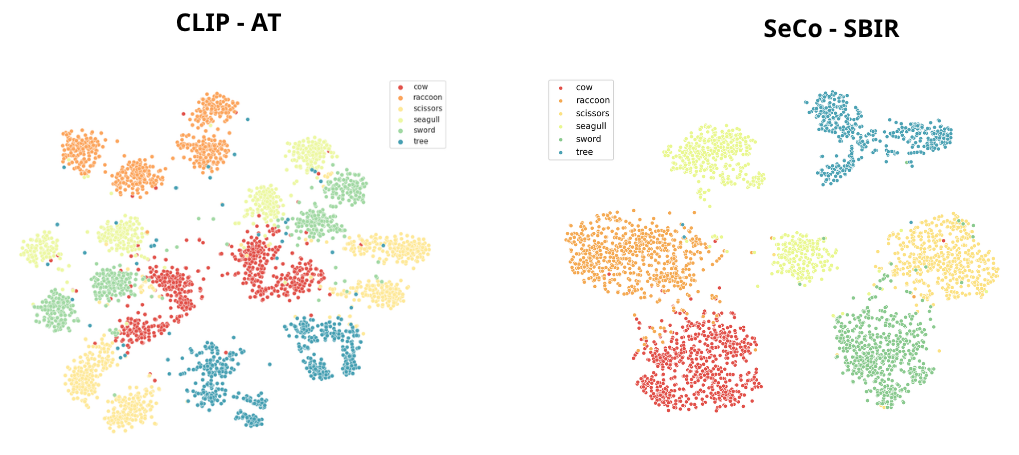}
    \caption{t-SNE visualization of sketch domain}
    \label{fig:tsne_sketch}
\end{figure*}

\noindent\textbf{Cross-domain alignment between sketches and photos.} We also illustrate t-SNE visualization for both photo and sketch modalities, compared between CLIP-AT method and SeCo-SBIR, in Figure \ref{fig:tsne_photo} and \ref{fig:tsne_sketch}, respectively. 

The two t-SNE plots show that SeCo-SBIR learns a more clearly separated feature space than CLIP-AT in both the photo domain and the sketch domain. In CLIP-AT, the clusters of several classes still overlap and are more scattered, suggesting that the features are not well separated across classes. In contrast, SeCo-SBIR forms more compact clusters with clearer distances between classes. 

This trend appears consistently in both domains, indicating that the proposed method not only improves inter-class discrimination but also makes intra
-class features more stable. These results suggest that SeCo-SBIR learns more effective visual representations than CLIP-AT for the SBIR task.

\begin{figure}[H]
    \centering
    \includegraphics[width=\linewidth]{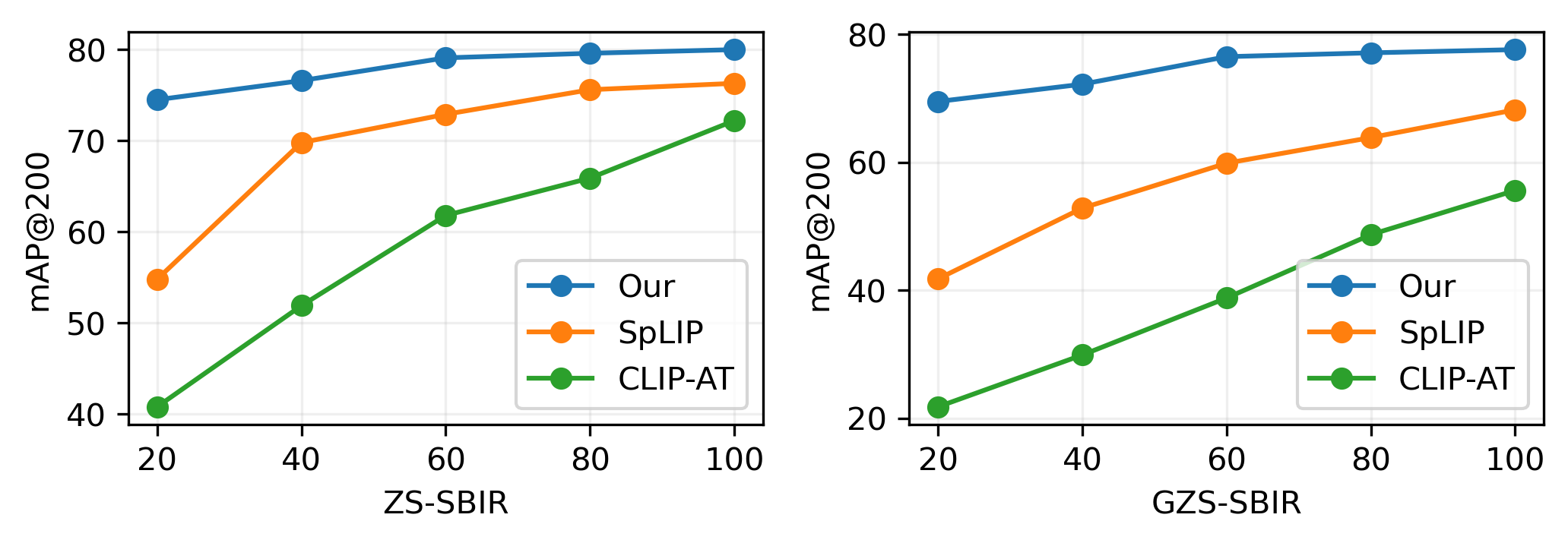}
    \caption{Comparison of SeCo-SBIR with previous methods with different training data size for Sketchy-Ext dataset}
    \label{fig:data_zs_gzs}
\end{figure}

\begin{figure}[H]
    \centering
    \includegraphics[width=\linewidth]{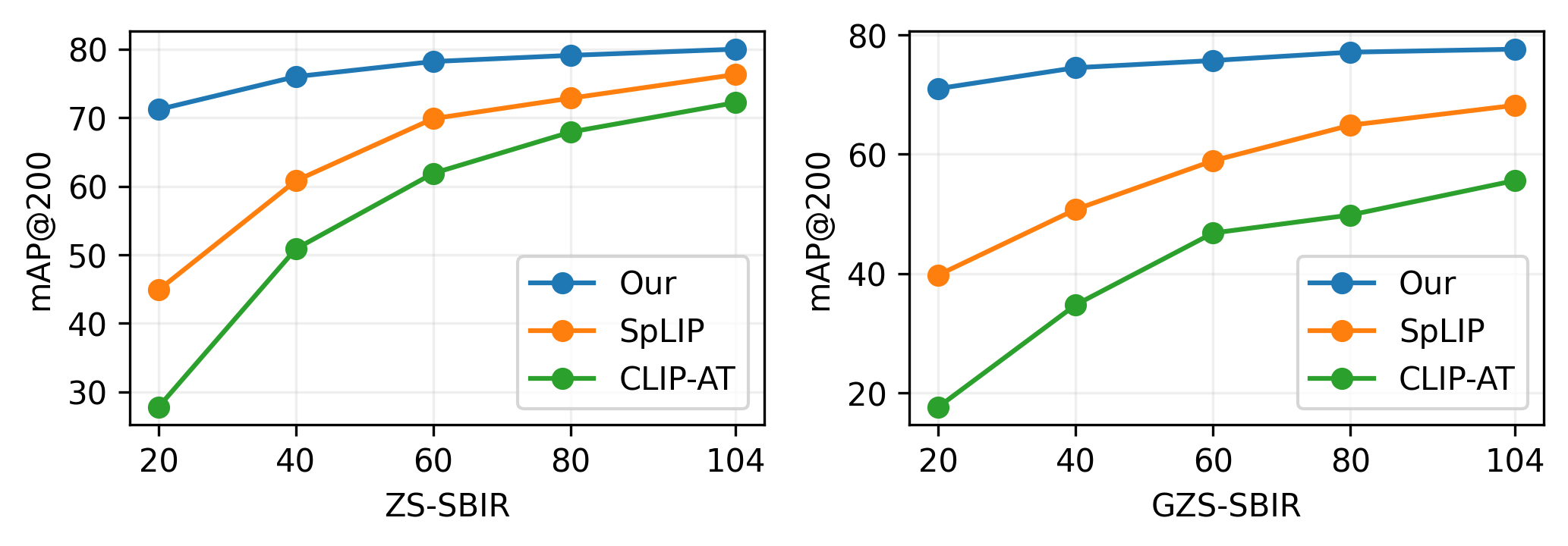}
    \caption{Comparison of SeCo-SBIR with previous methods with different numbers of seen classes for Sketchy-Ext dataset}
    \label{fig:classes_zs_gzs}
\end{figure}

\noindent\textbf{Generalization capability.} One of the main reasons for adopting CLIP is its strong behavior on out-of-distribution data which suggests good potential for deploying sketch-based applications at real-world scale. To further examine this property, we conduct experiments under two settings: (i) varying the amount of training data available for each class, using 20\%, 40\%, 60\%, 80\%, and 100\% (Figure \ref{fig:data_zs_gzs}); and (ii) changing the number of seen classes to 20, 40, 60, 80, and 104 on the Sketchy dataset (Figure \ref{fig:classes_zs_gzs}). The ZS-SBIR results are presented in the left-hand site of each image, while GZS-SBIR results are shown in the right-hand site.

Figure \ref{fig:data_zs_gzs} and \ref{fig:classes_zs_gzs} show that the performance of all methods improves as the training data size and number of classes increases. However, our method consistently achieves the best results on both ZS-SBIR and GZS-SBIR across all data scales and amount of seen classes. Notably, the gap between the proposed method and previous methods is more evident in the GZS-SBIR setting. This suggests that our model not only makes better use of the training data and seen classes, but also maintains stronger and more stable generalization ability than SpLip and CLIP-AT.


\end{document}